\documentclass[twoside,11pt]{article}

\usepackage{blindtext}

\usepackage{colortbl}  
\usepackage{algorithm}
\usepackage{booktabs}
\usepackage{amsmath}
\usepackage{algpseudocode}
\usepackage[abbrvbib, preprint]{jmlr2e}
\usepackage{jmlr2e}

\usepackage{lastpage}
\jmlrheading{..}{..}{1-\pageref{LastPage}}{..; Revised ..}{..}{21-0000}{JP Bruneton}

\ShortHeadings{Enhancing Symbolic Regression with Quality-Diversity and Physics-Inspired Constraints}{JP Bruneton}
\firstpageno{1}

\begin{document}

\title{Enhancing Symbolic Regression with Quality-Diversity and Physics-Inspired Constraints}

\author{\name Jean-Philippe Bruneton \email jpbruneton@gmail.com \\
       \addr  CNRS, LIED \\
       Université Paris Cité \\
       F-75013 Paris, France}

\maketitle

\begin{abstract}
This paper presents QDSR, an advanced symbolic Regression (SR) system that integrates genetic programming (GP), a quality-diversity (QD) algorithm, and a dimensional analysis (DA) engine. Our method focuses on exact symbolic recovery of known expressions from datasets, with a particular emphasis on the Feynman-AI benchmark. On this widely used collection of 117 physics equations, QDSR achieves an exact recovery rate of 91.6~$\%$, surpassing all previous SR methods by over 20 percentage points. Our method also exhibits strong robustness to noise. Beyond QD and DA, this high success rate results from a profitable trade-off between vocabulary expressiveness and search space size: we show that significantly expanding the vocabulary with precomputed meaningful variables (e.g., dimensionless combinations and well-chosen scalar products) often reduces equation complexity, ultimately leading to better performance. Ablation studies will also show that QD alone already outperforms the state-of-the-art. This suggests that a simple integration of QD, by projecting individuals onto a QD grid, can significantly boost performance in existing algorithms, without requiring major system overhauls.
\end{abstract}


\section{Introduction}

Symbolic Regression (SR) aims to discover mathematical expressions $y = f(\vec{x})$ that best describe a given dataset $(y_i, \vec{x}_i)$. 
While new supervised and unsupervised deep learning methods have emerged in recent years with promising results for symbolic regression tasks, see eg. \citep{kamienny_end--end_2022, tenachi_deep_2023},  genetic programming \citep{koza_genetic_1994} remains the method of choice. However, traditional GP is known to suffer in particular from premature convergence and bloat. The former arises because vanilla GP is inherently single-objective, aiming to produce increasingly well-adapted individuals with each generation.

In this paper, we build upon\footnote{\url{https://github.com/jpbruneton/QDSR}} a previously proposed GP-based model \citep{bruneton_exploration_2019}. This first model incorporated a projection of generated individuals onto a quality-diversity (QD) grid. Unlike traditional optimization techniques that aim to find a single optimal solution, QD algorithms explore the search space to illuminate a broad spectrum of behaviors or solutions. In practice, the MAP-Elites technique \citep{mouret_illuminating_2015} that we use here defines a QD-grid as a multi-dimensional array defined by user-selected phenotypes that can be extracted from candidate equations, such as equation length, the number of free parameters, or the number of variables used, etc. At each genetic iteration, candidate equations are thus evaluated and mapped to the appropriate cell in the QD grid based on their phenotypic features. If a new individual outperforms the one currently occupying a cell, it replaces the existing solution, ensuring continuous improvement while maintaining diversity. Crossovers are made between individuals living in the QD-grid, ie. having different phenotypes.

This approach inherently transforms GP into a multi-objective algorithm, as the grid retains the best individuals across a diverse range of features. This not only mitigates premature convergence and bloat, but also effectively maintains a high-dimensional Pareto front over key equation characteristics. As a result, it enhances interpretability by preserving a broad spectrum of solutions, allowing retrieval of optimal trade-offs between complexity and accuracy.

Also, by preserving high-performing short equations, the algorithm dynamically captures semantic 'building blocks' --- simple, interpretable subexpressions that are both syntactically compact and semantically meaningful. These building blocks can act as reusable components, akin to Automatically Defined Functions \citep{koza_genetic_1994} or Entropy-based 
Building-block Learning \citep{virgolin_scalable_2017}, to construct more complex and longer expressions in later stages of evolution. Furthermore, this aligns with techniques such as Grammar-Guided Genetic Programming (GGGP), where modular and semantically valuable structures are iteratively refined and reused \citep{whigham1995grammatically}.

In recent years, symbolic regression has become an increasingly active research area, expanding beyond computer science into phenomenological fields such as astrophysics, biology or health: \citep{wadekar_modeling_2020, matchev_analytical_2022, bartlett_exhaustive_2024, medina-ortiz_interpretable_2025, fong_symbolic_2024, abdussalam_symbolic_2025, tenachi_class_2024}. The Feynman AI dataset \citep{udrescu_ai_2020} has become a standard benchmark\footnote{See \url{https://space.mit.edu/home/tegmark/aifeynman.html}} for symbolic regression tasks, with many methods evaluated against it \citep{la_cava_contemporary_2021}. This dataset consists of 117 equations derived from Feynman lectures on physics and thus carry dimensional information as well. These equations are explicitly presented in Tables 3–6 below.

Compared to our previous work, the present model incorporates an expanded vocabulary along with on-the-fly dimensional analysis, as recommended in \citep{udrescu_ai_2020, tenachi_deep_2023}. As a result, we achieve up to 91.5 $\%$ strict exact recovery on the whole dataset, significantly outperforming the best-reported results to date, which, to the best of our knowledge, are around 70 $\%$ (see Table~\ref{tab:main_summary}). An ablation study (see section 4) reveals that our previous model was already superior to the current state of the art on this benchmark. However, it may have been overlooked or underappreciated due to the lack of published code, and its almost simultaneous release with the Feynman AI dataset. We thus rectify this by providing a thorough evaluation of the method, and open-sourcing our implementation.

\section{Related work and contributions}

\subsection*{Quality Diversity}
Quality Diversity algorithms are a subset of multi-objective evolutionary algorithms \citep{deb_fast_2002, deb_multi-objective_2004, cully_robots_2015}, designed not only to optimize performance but also to discover a diverse set of high-quality solutions. This approach has proven especially effective in fields like robotics, see for instance \citep{cully_robots_2015, pugh_quality_2016}.

The MAP-Elites algorithm of \citet{mouret_illuminating_2015} is a widely used QD method that partitions the search space into behavioral niches, optimizing within each niche independently to ensure diversity. Users define phenotypic features, or behavior descriptors, to characterize the niches based on problem-specific criteria, which are often orthogonal to performance metrics \citep{gaier_discovering_2020}. In symbolic regression, these features could include equation length, the number of variables, or free parameters.

To the best of our knowledge, this work is the first to investigate the combination of Quality Diversity and Genetic Programming for Symbolic Regression tasks. However, other multi-objective approaches have been applied to SR. For instance \citet{kommenda_genetic_2014} successfully addresses the issue of early convergence by evolving semi-independent subpopulations, enhancing diversity and mitigating premature convergence while exploring a broader part of the solution space. This is implemented in particular in the open-source PySR library of \citet{cranmer_interpretable_2023}. Also, Age-Fitness-Pareto optimization \citep{schmidt_age-fitness_2010} balances exploration and exploitation by maintaining individuals of varying ages and fitness; see also \citep{liu2022evolvability}. These approaches share similarities with ours in that they introduce a form of multidimensional or partitioned genetic pool, promoting diversity and preventing premature convergence.

\subsection*{Symbolic regression}
Recent years have seen rapid growth in SR approaches \citep{angelis_artificial_2023}. They can broadly be categorized into GP-based methods \citep{virgolin_scalable_2017, virgolin_linear_2019, tohme_gsr_2022}, deep learning methods using reinforcement or supervised learning \citep{martius_extrapolation_2016, biggio_neural_2021, valipour_symbolicgpt_2021, kim_integration_2021, vastl_symformer_2024, kamienny_end--end_2022, tenachi_deep_2023, tian_interactive_2025}, or hybrid approaches combining these techniques \citep{cranmer_discovering_2020, mundhenk_symbolic_2021, petersen_deep_2020, landajuela_unified_2022, holt_deep_2023}. Some other models restrict themselves to parametrized solution space \citep{kammerer_symbolic_2022, scholl_parfam_2024}. Emerging methods are also beginning to explore zero-shot learning capabilities of large language models (LLMs) to tackle SR tasks \citep{grayeli_symbolic_2024, li_mmsr_2025}. For more comprehensive reviews, see \citep{la_cava_contemporary_2021, angelis_artificial_2023, makke_interpretable_2024, radwan_comparison_2024}.

The Feynman AI dataset is a widely recognized benchmark in symbolic regression \citep{udrescu_ai_2020}, yet it has been criticized for including overly simplistic targets, redundant equations, and impractical variable ranges. To address these shortcomings, recent initiatives, such as the introduction of the SRSD dataset \citep{matsubara2024}, have refined the benchmark by curating its content and establishing more realistic variable ranges. We note that variables can always be whitened, meaning that SRSD with whitened variables is essentially the Feynman-AI dataset with transformations of the form \( x_i \to \alpha_i x_i + \beta_i \), effectively increasing the difficulty. Due to computational limitations, we have not yet conducted experiments on this dataset.

Dimensional analysis has proven to be an important tool in SR, particularly for physics-based datasets like the Feynman one. Methods such as AI Feynman \citep{udrescu_ai_2020, udrescu_ai_2020-1} and more recent advancements \citep{tenachi_deep_2023} incorporate dimensional constraints to enhance exact recovery. As it turns out, 26 out of the 117 equations of the dataset are dimensionally trivial and can therefore be solved exactly in constant time, establishing a baseline exact recovery rate of $26/117 \sim 22 \%$.

An analysis of the existing literature reveals that the presented models are often restricted to a relatively short list of elementary functions, such as (sin, exp, log). Consequently, they exclude target expressions that involve functions like arcsin from the outset \citep{la_cava_contemporary_2021}. While it may seem reasonable to initially reduce the size of the search space, excluding such targets is problematic, as it effectively assumes prior knowledge about the underlying expressions—an assumption that should not be made. Instead, we include a comprehensive set of functions, encompassing all trigonometric and hyperbolic functions along with their inverses.

Finally, high-performance SR tools such as PySR and Operon \citep{cranmer_interpretable_2023, burlacu_operon_2020} have been developed to accelerate SR through efficient algorithms and parallelism. Our model is at the contrary not optimized for high performance; this is left for future work.

\section{Description of the model}

\subsection{Main algorithm}
\begin{algorithm}
\caption{Overview of Quality-Diversity Symbolic Regression (QDSR)}
\begin{algorithmic}[1]
    \State Initialize random population $P$ satisfying DA constraints
    \State Initialize an empty QD grid
    \While{budget $<$ max\_budget}
        \State Subsample 250 data points 
        \State Evaluate the fitness of each equation in $P$
        \If{stopping criteria met}
            \State \textbf{break}
        \EndIf
        \State Extract features from each equation in $P$ 
        \State Update QD grid: retain only the best individual per feature vector
        \State Select parent equations from the QD grid
        \State Apply DA-consistent genetic operators (crossover, mutation) to generate offspring
        \State $P \gets$ offspring
    \EndWhile
    \State \Return best discovered equations from the QD grid
\end{algorithmic}
\end{algorithm}
The above pseudo-code provides a high-level overview of the main algorithm. Before delving into the details of its individual components, we briefly outline its key steps. The initialization step (line 1) requires three main components: a dimensional analysis engine to enforce physical consistency, a tree-based representation of symbolic expressions and its associated vocabulary, and a random tree generator to create diverse initial expressions.

Next, explicit features must be defined to characterize candidate equations (line 2). In our case, we use the feature vector: $\textrm{len(eq)}$, $\textrm{num}(\textrm{free scalars})$, $\textrm{num(functions)}$, and $\textrm{num(variables)}$, see also Fig. \ref{dimtree} for an explicit example. The maximum budget (line 3) is typically constrained by factors such as time limits, the maximum number of iterations, or the total number of unique individuals evaluated. These parameters are further discussed in Section 4.

Subsampling (line 4) is a standard practice in symbolic regression, as datasets like Feynman AI contain up to $10^6$ data points, making free scalar optimization prohibitively expensive. Plus, in real-world applications, experimental data would typically be far less abundant. Prior work \citep{scholl_parfam_2024} for instance used 500 data points. We have found that values between 200 and 1000 generally yield good results. Fitness evaluation in line 5 follows a two-step process: first, optimizing the free scalars via least squares fitting, then computing the fitness score, typically using $R^2$ or NRMSE (see below). The early stopping criteria (lines 6–8) are based on this fitness evaluation.

As a result, the QD grid is updated by clustering individuals with the same feature vector (line 10). Among them, only the fittest is retained. This new candidate is compared to the existing representative in the QD grid and replaces it if it has a better fitness score. Selection for breeding follows (line 11). We use a strategy that slightly biases selection toward high-performing individuals of the QD grid. 

Finally, each iteration generates a new population of the same size as the current QD grid (lines 12–13), and the process repeats.

\subsection{Dimensional Motor}
Given units of variables and of the target, straightforward linear algebra and implementation with SymPy enable us to determine whether an equation is dimensionally trivial. Additionally, this engine groups variables with shared dimensions (if any) and identifies possible dimensionless combinations. These capabilities play an important role in the vocabulary definition, see below.

\subsection{Tree representation}
\begin{figure}[ht]
    \centering
    \includegraphics[width=0.8\linewidth]{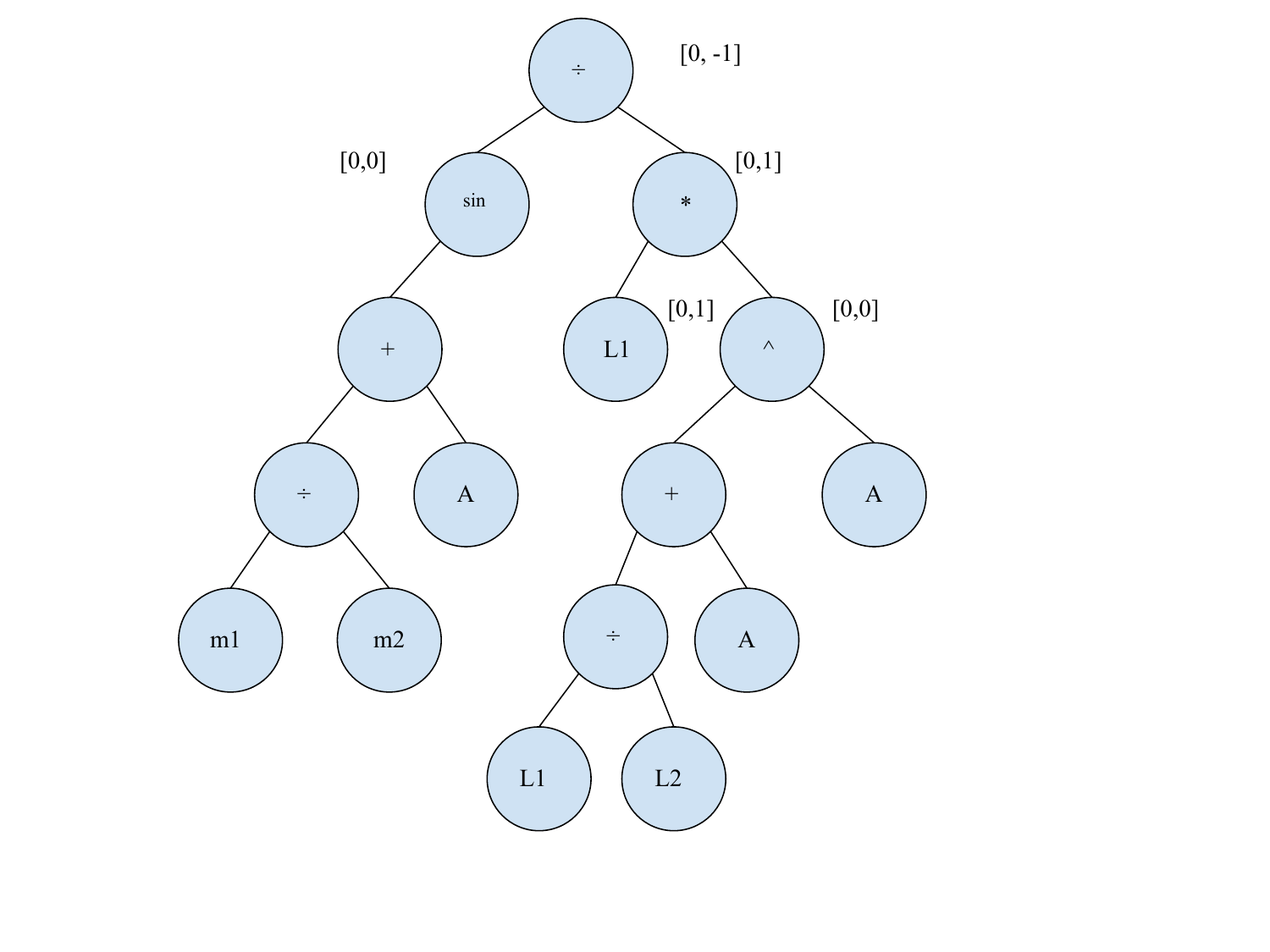}
    \caption{The tree representing the equation $\left(\sin(\frac{m_1}{m_2} + A)\right)/\left(L_1*\left( A + \frac{L_1}{L_2}\right)^A\right)$ for two masses $(m_1, m_2)$ and two lengths $(L_1, L_2)$; in front of some nodes we have shown the dimension they carry, in the format $[M, L]$ where $L$ stands for Length and $M$ for mass. The resulting dimension is $[0, -1]$. This equation has the following feature vector (16, 3, 1, 5): 16 nodes, 3 free scalars, one unary function, and 5 occurrences of variables. The three 'A's should be understood as distinct but yet undetermined scalars: $(A_1, A_2, A_3)$.}
    \label{dimtree}
\end{figure}

As is standard in tree-based genetic programming, candidate equations are represented as abstract syntax trees, with the added feature that each node carries both a symbol and its corresponding dimension (e.g., a length). This approach is well-documented in the literature, see e.g. \citep{makke_interpretable_2024}, so we do not provide further details here. Fig. \ref{dimtree} illustrates how dimensions propagate through the tree. 
In our implementation, nodes are further augmented with metadata that allows us to enforce another constraint, namely limiting the number of nested functions and powers to a predefined value, $K$. For this work, we have set $K = 2$ as a hard limit. It is important to note that nested functions can sometimes provide an advantage. For example, a target like I.48.2: $E_n = m c^2/\sqrt{1-v^2/c^2}$ might be challenging to discover directly because of the denominator of the form $\sqrt{1-x^2}$. However, this denominator can alternatively be expressed as $\sin(\arccos(x))$ or even found through  $\cos(\arctan(x)) = x/\sqrt{1 -x^2}$, etc. By leveraging such alternative formulations, the algorithm often finds the correct solution with fewer symbols, demonstrating the strategic advantage of using a rich and expressive vocabulary.

\subsection{Vocabulary}
The vocabulary used in our runs is structured as follows:

\begin{itemize}
\item \textbf{Binary operations}: $(+, -, *, \div, **)$ are included. The exponentiation operation $f(x)^{g(x)}$ is permitted if both $f$ and $g$ are dimensionless; otherwise, the exponent must be an integer such that the resulting dimension matches the node's dimension.
\item \textbf{Unary functions}: The unary functions includes $(\exp, \log, \textrm{sqrt})$, along with all trigonometric functions: $(\sin, \, \cos, \, \tan, \, \arcsin,\,  \arccos,\,  \arctan)$ as well as $(\sinh,\,  \cosh, \, \tanh)$ and $(\textrm{arcsinh},\,  \textrm{arccosh},\,  \textrm{arctanh})$. Unary functions can only be applied to dimensionless quantities, with the exception of the square root.
\item \textbf{Leaves}: Leaf nodes can be either $A$ or $x_i$, where $A$ represents a placeholder for free scalar parameters to be optimized later, and $x_i$ denotes the original input variables. Leaf nodes may also occasionally be integers, when the parent node represents a power --- for instance, $x^2$, for a node with the dimension of a surface, where $x$ has the dimension of a length.
\end{itemize}

However, dimensional analysis allows us to \textbf{expand the leaves vocabulary} to include dimensionless combinations, or scalar products of variables sharing the same dimension. In order to clarify this procedure, we present an example involving four length-based variables, corresponding to target I.8.14, where the target function is given by $f = \sqrt{(x_2-x_1)^2+(y_2-y_1)^2}$. We first rename the variables $(x_0, x_1, x_2, x_3)$. In this case, the following additional elements are included in the leaf vocabulary:
\begin{itemize}
\item \textbf{Dimensionless variables}: We introduce the ratios $y_0 = \frac{x_0}{x_1}, y_1 = \frac{x_0}{x_2}, y_2 = \frac{x_0}{x_3}$, along with their respective reciprocals. Additionally, an optional setting\footnote{Set to true in the main run.} allows for the inclusion of all possible products between the $y_i$.
\item \textbf{Norms}: We add the sum of all squares of variables sharing a common dimension. In this case, it is $n_0 = \sum_i x_i^2$. 
\item \textbf{Scalar products}: We add expressions of the form $n_1 = (x_0 - x_1)^2 +  (x_2 - x_3)^2$, along with all other non-equivalent permutations $n_j$. The number of such unique scalar products corresponds to the number of possible pairings, given by: $\frac{n!}{2^{n/2} (n/2)!}$
\end{itemize}
For example, this results in 3 unique pairings for $n = 4$ and 15 pairings for $n = 6$, as seen in target I.9.18, given by 
\begin{equation}
    f = \frac{G m_1 m_2}{((x_2-x_1)^2+(y_2-y_1)^2+(z_2-z_1)^2}.
    \label{1918}
\end{equation}

Adding these variables to the vocabulary is justified for several reasons. First, dimensionless ratios typically appear as arguments of unary functions and may be difficult to discover independently. A clear example is target I.40.1, where the true formula is $f \propto \exp(-m g x/(k_B T))$. Introducing the $y$ variables into the vocabulary allows the equation to be rewritten in a much simpler way as $f \propto exp(-y_j)$ for one of the dimensionless variables $y_j$. 

The remaining benefits align with some ideas behind the AI-Feynman engine of \citet{udrescu_ai_2020}, which exploits physical properties inherent in data distributions. Adding the norm $n_0$ for instance facilitates the discovery of equations that exhibit spherical symmetry in the space of these variables, while including scalar products (particularly for coordinates with length dimensions) effectively incorporates translation invariance, and computes all relevant physical distances needed for the regression engine.

Naturally, we immediately see the advantage of incorporating these physics-inspired terms. Although this significantly increases the vocabulary size, it has the benefit of making the target equation much easier to discover. For instance, target I.8.14 becomes almost trivial since it simplifies to $f = \sqrt{n_j}$ for one of the norms $n_j$ introduced in the vocabulary. Similarly, target I.9.18 in Eq.~(\ref{1918}), which would otherwise be extremely difficult to recover, becomes much more accessible. Although we introduce 50 auxiliary variables in addition to the original 9 variables, the target equation simplifies to 
$$
f = \frac{G m_1 m_2}{n_j}
$$  
for one of the scalar products $n_j$. Given that dimensional analysis further constrains the functional form to
$$
G \cdot \frac{\mathrm{mass}^2}{\mathrm{length}^2},
$$  
the equation becomes so trivial that it is frequently present in the initial random pool. This highlights a key trade-off: although dramatically increasing the search space size may seem counterproductive, it is beneficial if it significantly simplifies the discovery of the target equation.

\subsection{Dimensionally Consistent Random Tree Generator}  
Generating an initial random pool of candidate equations while enforcing multiple constraints is a challenging task. These constraints include limiting equation length (35 or 45, see below), controlling function nesting, ensuring dimensional consistency, and maintaining valid dimensional relationships throughout. We developed a solution, detailed in the appendix, that yield diverse enough trees though we do not claim it is optimal. In practice, we generate 10,000 random individuals at the beginning of the run, yielding around 8,500 unique trees after removing duplicates.  

\subsection{Quality-Diversity Grid}  
As mentioned earlier, our runs utilize a four-dimensional QD grid. For each candidate equation, we extract four phenotypic characteristics: equation length (measured by the number of nodes in its tree representation), the number of free scalar parameters $A$, the number of variable occurrences in the formula, and the number of unary functions. We use a bin size of 1 for the first two characteristics and a bin size of 2 for the latter two. With a maximum equation length set to 35, the QD grid typically has 1,000 individuals.

One detail previously omitted in Section 3.1 concerns how we handle long-running searches. If the target equation has not been found within a one-hour timeout, we project the current grid onto a larger one that introduces a fifth dimension --- a boolean indicating whether the formula contains nested functions. Simultaneously, we increase the maximum equation length by 10, allowing formulas up to a length of 45. Under these settings, the typical QD grid expands to around 4,500 elements.

\subsection{Genetic operations}
The genetic operations used in our approach follow standard principles but incorporate constraints to ensure dimensional consistency.  
\begin{itemize}
\item Crossover: Allowed only between nodes of the same physical dimension to maintain valid dimensional relationships in offspring equations.  
\item Mutation: Node mutation must also respect dimensional analysis.
\item Unary Removal: Unary functions tend to accumulate excessively, particularly due to the allowance of nested functions. To prevent unnecessary complexity and reduce the false positive rate, we periodically remove unary nodes at random.
\item Parent Selection: Parents are selected from the flattened QD grid, with a preference for high-performing individuals. Selection follows a \( 1/(x + k) \) probability distribution where $x$ is the rank and $k$ some constant, favoring better-performing candidates while still allowing diversity in the selection process.  
\end{itemize}
At each iteration, a new population is generated with a size equal to the current QD grid. Approximately one-third of the new elements are produced through crossovers, one-third through mutations, and one-third through unary removals.

\subsection{Free scalar determination}
Given data points $(x_i, y)$ and a candidate equation $f(x_i, A_j)$,  the optimal values for the free scalar parameters are determined by minimizing the mean squared error: $(y - f(x_i, A_j))^2$. To improve computational efficiency, we use SciPy’s least squares optimizer instead of the more commonly used BFGS. Another alternative would be a global optimizer such as CMA-ES \citep{hansen_cma-es_2011}, which is highly effective but orders of magnitude slower. As explained in Section 3.1, at each iteration, we draw a new subsample of 250 data points from the training dataset. This strikes a good balance between computational cost and optimization accuracy.

\subsection{Fitness; early stopping criteria; false positives}
Both the Normalized Root Mean Square Error (NRMSE) and the coefficient of determination (\( R^2 \)) can be used to evaluate the fitness of individuals. We use the latter in the runs. The formulas are as follows for a batch of $N$ data points :  
\begin{equation}
\mathrm{NRMSE} = \frac{\sqrt{\frac{1}{N} \sum_{i=1}^{N} (y_i - \hat{y}_i)^2}}{\sigma_y}
\end{equation}
where \( y_i \) are the predicted values, \( \hat{y}_i \) are the true data points, and \( \sigma_y \) is the standard deviation of \( \hat{y} \).  
\begin{equation}
R^2 = 1 - \frac{\sum (y_i - \hat{y}_i)^2}{\sum (\hat{y}_i - \bar{y})^2}
\end{equation}
where \( \bar{y} \) denotes the mean of the true data \( \hat{y}_i \). Since our focus is on exact recovery---rather than approximate recovery up to an additive or multiplicative constant, as in \citet{la_cava_contemporary_2021}---we adopt a strict early stopping criterion of \( R² \geq 1 - 10^{-14} \), close to numerical precision.

To determine whether an equation matches the ground truth exactly, we first attempt simplification using SymPy. This involves substituting the optimized values of free scalar parameters and rounding them to the nearest integer or to 'physics-like numerical forms'. These forms, inspired by common physical and geometrical constants, ensure that numerical values are clipped to meaningful approximations.\footnote{Specifically, we consider numbers of the form $r \sqrt{n} \pi^q$ for some predetermined list of rationals $r$, integers n, and exponents $q$. For instance it allows to recognize the number $0.6266570686577501$ as being $\sqrt{2 \pi}/4$.} If a numerical value is within $10^{-9}$ of such values,  it is rounded accordingly. In most cases, SymPy then correctly detects exact matches, but when it fails, we verify the equivalence manually or using Wolfram Alpha.  

The early stopping criterion is only applied to the current subsampled batch and not on the whole dataset. If it is met, the genetic process terminates, and the candidate equation is evaluated for exact recovery. 

In some cases, exact recovery is not achieved despite an $R^2$ score of 1 up to numerical precision. These rare occurrences typically fall into two categories. The first involves candidate equations that closely resemble the ground truth but contain additional terms with large coefficients, such as expressions of the form $1 - \tanh(A x)$ with a large $A \sim O(1000)$, which can be numerically indistinguishable from zero. The second category consists of equations in the dataset that are particularly prone to false positives. This is especially true for the first five rows of Table \ref{tab:hard}, where both our method and PySR tend to converge prematurely to highly accurate approximations of the target function, such as high-order Taylor series expansions.

\subsection{Train and test sets}
For exact symbolic recovery, splitting the dataset into training and test sets is unnecessary. However, we included a 75-25 \% split in the code for generality. Note that each genetic iteration operates on a randomly subsampled batch of 250 data points. Since none of our runs exceeded 300 iterations, the total number of data points used in any run was at most 75,000, out of the $10^6$ available in the Feynman AI dataset. This means that, in practice, we trained on less than 10\% of the full dataset.

\subsection{Handling Noisy Data}
Following \citet{la_cava_contemporary_2021}, we also evaluate the regressor on manually perturbed data. Given the original target values \( f \), we introduce additive Gaussian noise according to the formula:
\begin{equation}
\sigma_f = \textrm{std}(f), \quad f_{\textrm{noise}} = f + \eta, \quad \eta \sim \mathcal{N}(0, \sigma_f \cdot \textrm{noise\_level}).
\end{equation}
The treatment of exact symbolic recovery in the presence of noise is sometimes ambiguous in the literature. In principle, it becomes impossible as soon as free scalar parameters are involved. For example, in an expression such as \( \cos(Ax) \), the optimal value of \( A \) will inevitably differ between noisy and noiseless data.  

To address this issue, we adopt the approach of \citet{tenachi_deep_2023}. The QD grid is evolved using only noisy data; however, at each genetic iteration, we extract a complexity-fitness Pareto front consisting of approximately 50–70 candidate equations. These individuals are then refitted against the original noiseless data, and early stopping decisions are made based on this refined evaluation. The complexity of an equation is computed as the sum of token-wise complexity scores, assigned as follows:
\begin{table}[ht]
    \centering
    \begin{tabular}{ll}
        \toprule
        \textbf{Operation} & \textbf{Complexity Score} \\
        \midrule
        Input Variables & 1 \\
        Basic arithmetic & 1: \( (+, *) \), 2: \( (-, \div) \), 3 : \( (**) \) \\
        Free scalar parameters & 2 \\
        Exponential/log functions & \( \log, \exp, \sqrt{\, \, } \): 3 \\
        Trigonometric functions & \( \sin, \cos, \tan \): 4 \\
        Hyperbolic functions & \( \sinh, \cosh, \tanh \): 5 \\
        Inverse trigonometric functions & \( \arcsin, \arccos, \arctan \): 5 \\
        Inverse hyperbolic functions & \( \textrm{arcsinh}, \textrm{arccosh}, \textrm{arctanh} \): 6 \\
        \bottomrule
    \end{tabular}
    \caption{Complexity scores assigned to different mathematical operations.}
    \label{tab:cscore}
\end{table}

\section{Runs and Results}  
We conducted five types of runs: two main runs (one on noiseless data and one with a noise level of 0.1), two ablation studies on noiseless data, and a reference PySR run using the same set of unary functions as our method. 

\subsection{Main Run}  
The 'main run' refers to 10 independent runs conducted on the entire dataset, with dimensional analysis and additional variables enabled as described earlier. When running with 40 threads, we set a timeout of 1 hour using the 'medium-sized' QD grid described in Section 3.5. If the target equation is not found within this period, the run continues for up to an additional 5 hours, increasing the maximum expression length to 45 and switching to the larger QD grid (also described in Section 3.5). In all cases, evolution stops once \(10^6\) individuals have been evaluated, following the stopping criterion of \citet{la_cava_contemporary_2021}. However, in practice, runs typically terminate after approximately 750,000 evaluations within 6 hours. 

\subsection{Ablation Studies}  
To assess the impact of different components, we conducted two ablation studies:
\begin{itemize}
\item \textbf{Run Ab\_0}: This run disables dimensional analysis entirely (by setting all dimensions to zero) and excludes additional variables. 
\item \textbf{Run Ab\_1}: The system runs with DA enabled but without any additional variables. 
\end{itemize}
All other settings remain unchanged. To save computational resources, these ablation runs are averaged over 5 independent runs instead of 10.

\subsection{Reference run : PySR}
We ran PySR with a fairly standard configuration, using the same list of unary functions but without allowing nested functions, as unary removal cannot be implemented in PySR. We used 15 populations, each with a size of 100, and 500 cycles per iteration, with both maxsize and maxdepth set to 35. The run was limited to a 30-minute timeout on 40 processors, resulting in significantly more individuals being evaluated compared to our runs, thanks to the impressive speed of PySR's backend.

\subsection{Results}
Next Table \ref{tab:main_summary} is the high-level view of our results. It shows the average of the exact symbolic recovery over the runs and the 117 targets. We ran PySR five times and obtained results consistent with those reported in the literature \citep{scholl_parfam_2024}. uDSR results were taken from \citep{landajuela_unified_2022, tenachi_deep_2023}, Phy-SO results from \citep{tenachi_deep_2023}, and AI-Feynman 2.0 (AIF-2) results from \citep{la_cava_contemporary_2021}. Methods with a recovery rate below 40\% have been excluded from the table.

\renewcommand{\arraystretch}{1.25} 
\begin{table}[ht]
\centering
\begin{tabular}{|l|c|}
\hline
\multicolumn{2}{|c|}{\textbf{Noiseless Feynman AI DataSet Benchmark}} \\ \hline
Method & Exact Symbolic Recovery Rate\\ \hline
\rowcolor[gray]{0.9} QDSR (Main run) & 91.6\%  \\ \hline
\rowcolor[gray]{0.9} QDSR (\textrm{Ab\_0} run)  & 74.2\%  \\ \hline
\rowcolor[gray]{0.9} QDSR (\textrm{Ab\_0} run)  & 85.1\%  \\ \hline
\rowcolor[gray]{0.9} QDSR (Main run with 0.1 noise) & 79.4\% \\ \hline
uDSR & $\approx  70 \%$ \\ \hline
PySR  & 65.8\%  \\ \hline
PhySO & 58.5\%  \\ \hline
ParFam & $\approx 58 \%$ \\ \hline
AIF-2 & $\approx 53\%$ \\ \hline
\end{tabular}
\vspace{0.5cm}
\caption{Total exact recovery rate of various methods on the noiseless dataset (except for line four), comparing our approach to the best methods from the current literature.}
\label{tab:main_summary}
\end{table}
It is interesting to note that QDSR alone (Ab\_0 run), without dimensional analysis or an enhanced leaf vocabulary, outperforms state-of-the-art methods, achieving an exact recovery rate of 74\%. This observation may be of interest to other researchers, as it suggests that a simple modification to existing algorithms --- by implementing, maintaining, and projecting individuals onto a QD grid --- can lead to significantly better performance without needing to overhaul the entire system.

Dimensional analysis improves performance by 10 percentage points, while introducing extraneous variables allows us to solve otherwise difficult targets, such as I.9.18 in Eq.~(\ref{1918}), without compromising the ability to solve other equations. Note that this approach can also be easily implemented in any method, including those that are not GP-based. In the end, we achieve nearly 92\% exact recovery.

\subsection{Detailed results}
For completeness, we present the detailed results by target. First, the 26 dimensionally trivial targets in the benchmark are solved in constant time using dimensional analysis. These targets must be monomials of the form $f = A x_1^{q_1} \ldots x_n^{q_n}$, and are in fact easily solved as well even without DA. Both PySR and run Ab\_0 also achieve a $100 \%$ hit rate on these targets.

Beyond the dimensionally trivial cases, there are 22 additional targets for which both PySR and our method achieve $100\%$ symbolic recovery. These are simple targets, such as almost monomials like II.37.1 : $f = \textrm{mom} (1+\chi) B$, or II.6.11: $f = p_d \cos(\theta)/(4\pi \varepsilon r^2)$,  or even simple enough equations like target I.11.19: $f = x_1 y_1+x_2 y_2+x_3 y_3$.\footnote{The complete list is : I.11.19, I.12.1, I.12.2, I.13.12, I.18.12, 
I.18.14, I.39.22, II.6.11, II.10.9, II.11.20, 
II.15.4, II.15.5, II.37.1, II.38.14, III.12.43, 
III.13.18, III.15.27, I.27.6, II.38.3,  
I.12.11, II.2.42, III.15.12}

The remaining equations are presented in the following tables. We categorize the set into 35 easy targets (for our method), shown in Tables \ref{tab:easy1} and \ref{tab:easy2}, 20 medium difficulty targets in Table \ref{tab:medium}, and the hard targets in Table \ref{tab:hard}.

\newpage
\begin{table}[ht]
    \centering
    \begin{tabular}{|l|c|c|>{\columncolor[gray]{0.9}}c|c|c|c|}
    \hline
    \multicolumn{7}{|c|}{\textbf{35 Easy Targets}} \\ 
    \hline
    \multicolumn{2}{|c|}{\textbf{Equation Details}} & \multicolumn{5}{c|}{\textbf{Hit Rates in \%}} \\
    \hline
    \textbf{Eq. Id.} & \textbf{Formula} & \textbf{PySR} & \textbf{Main} & \textbf{Abl 0} & \textbf{Abl 1} & \textbf{w. noise} \\
    \hline
        $\mathrm{B}_5$ & $\frac{2 \pi d^{3/2}}{\sqrt{G (m_1 + m_2)}}$ & 100 & 100 \%  & 80 & 100 & 60\\
        $\mathrm{B}_{18}$ & $\frac{3}{8 \pi G} \left( \frac{c^2 k_f}{r^2} + H_G^2 \right)$ & 100 & 100 \%  & 100 & 100 & 100 \\
        I.34.14 & $\frac{(1 + v/c)}{\sqrt{1 - v^2 / c^2}} \omega_0$ & 20 & 100 \%  & 80 & 100& 100\\
        II.11.3 & $\frac{q E_f}{m (\omega_0^2 - \omega^2)}$ & 80 & 100 \%  & 100 & 100& 100\\
        I.43.43 & $\frac{1}{\gamma - 1} \frac{k_b v}{A}$ & 100 & 100 \%  & 80 & 100& 100 \\
        I.47.23 & $\sqrt{\gamma p_r / \rho}$ & 100 & 100 \%  & 80 & 100& 100\\
        I.34.1 & $\frac{\omega_0}{1 - v/c}$ & 100 & 100 \%  & 80 & 100& 100\\
        I.39.11 & $\frac{1}{\gamma - 1} p_r V$ & 100 & 100 \%  & 60 & 100& 100 \\
        III.19.51 & $-\frac{m q^4}{2 (4 \pi \varepsilon)^2 (h / (2 \pi))^2} \frac{1}{n^2}$ & 80 & 100 \%  & 80 & 100& 100\\
        II.34.11 & $\frac{g_* q B}{2 m}$ & 100 & 100 \%  & 80 & 100& 100\\
        II.34.29b & $\frac{g_* \mathrm{mom} B J_z}{h / (2 \pi)}$ & 100 & 100 \%  & 80 & 100& 100\\
        I.40.1 & $n_0 \exp\left(-\frac{m g x}{k_b T}\right)$ & 40 & 100 \%  & 100 & 100 & 100\\
        II.13.23 & $\frac{\rho_{c_0}}{\sqrt{1 - v^2 / c^2}}$ & 0 & 100 \%  & 80 & 80 & 80\\
        II.13.34 & $\frac{\rho_{c_0} v}{\sqrt{1 - v^2 / c^2}}$ & 0 & 100 \%  & 40 & 100 & 80\\
        I.10.7 & $\frac{m_0}{\sqrt{1 - v^2 / c^2}}$ & 20 & 100 \%  & 40 & 100 & 80 \\
        II.6.15b & $\frac{p_d}{4 \pi \varepsilon} \frac{3 \cos(\theta) \sin(\theta)}{r^3}$ & 80 & 100 \%  & 100 & 100 & 100\\
        II.11.27 & $\frac{n \alpha}{1 - (n \alpha / 3)} \varepsilon E_f$ & 80 & 100 \%  & 100 & 100 & 100\\
        II.36.38 & $\frac{\mathrm{mom} H}{k_b T} + \frac{\mathrm{mom} \alpha}{\varepsilon c^2 k_b T} M$ & 100 & 100 \%  & 60 & 20 & 100\\
        I.50.26 & $x_1 (\cos(\omega t) + \alpha \cos(\omega t)^2)$ & 40 & 100 \%  & 100 & 100 & 80\\
        III.17.37 & $\beta (1 + \alpha \cos \theta)$ & 100 & 100 \%  & 80 & 80 & 100\\
        I.18.4 & $\frac{m_1 r_1 + m_2 r_2}{m_1 + m_2}$ & 100 & 100 \%  & 100 & 100 & 100\\
        $\mathrm{B}_9$ & $-\frac{32}{5} \frac{G^4}{c^5} (m_1 m_2)^2 \frac{(m_1 + m_2)}{r^5}$ & 0 & 100 \%  & 80 & 80 & 60\\
        I.8.14 & $\sqrt{(x_2 - x_1)^2 + (y_2 - y_1)^2}$ & 40 & 100 \%  & 0 & 100 & 100\\
        I.13.4 & $\frac{1}{2} m (v^2 + u^2 + w^2)$ & 60 & 100 \%  & 100 & 100& 100\\
        I.24.6 & $\frac{1}{2} m (\omega^2 + \omega_0^2) \frac{1}{2} x^2$ & 40 & 100 \%  & 100 & 100& 100 \\
        \hline
    \end{tabular}
    \caption{25 very easy targets; continued on the next page. B refers to the 'bonus' equation list of SR-Bench; see \url{https://space.mit.edu/home/tegmark/aifeynman.html}}
    \label{tab:easy1}
    \end{table}
\begin{table}[ht]
    \centering
    \begin{tabular}{|l|c|c|>{\columncolor[gray]{0.9}}c|c|c|c|}
    \hline
    \multicolumn{7}{|c|}{\textbf{35 Easy Targets (continued)}} \\ 
    \hline
    \multicolumn{2}{|c|}{\textbf{Equation Details}} & \multicolumn{5}{c|}{\textbf{Hit Rates in \%}} \\
    \hline
    \textbf{Eq. Id.} & \textbf{Formula} & \textbf{PySR} & \textbf{Main} & \textbf{Abl 0} & \textbf{Abl 1} & \textbf{w. noise} \\
    \hline
    II.35.21 & $n_{\rho} \mathrm{mom} \tanh(\mathrm{mom} B / (k_b T))$ & 60 & 100 \%  & 80 & 80& 100\\
    III.10.19 & $\mathrm{mom} \sqrt{B_x^2 + B_y^2 + B_z^2}$ & 0 & 100 \%  & 0 & 60& 100\\
    I.9.18 & $\frac{G m_1 m_2}{(x_2 - x_1)^2 + (y_2 - y_1)^2 + (z_2 - z_1)^2}$ & 0 & 100 \%  & 0 & 0& 100\\
    I.44.4 & $n k_b T \ln(V_2 / V_1)$ & 20 & 100 \%  & 80 & 80& 100 \\
    II.21.32 & $\frac{q}{4 \pi \varepsilon r (1 - v/c)}$ & 80 & 100 \%  & 80 & 100& 100 \\
    III.14.14 & $I_0 (\exp(q \mathrm{Volt} / (k_b T)) - 1)$ & 80 & 100 \%  & 80 & 100& 100\\
    I.32.17 & $\left(\frac{1}{2} \varepsilon c E_f^2\right) \left(\frac{8 \pi r^2}{3}\right) \left(\frac{\omega^4}{(\omega^2 - \omega_0^2)^2}\right)$ & 0 & 100 \% & 80 & 100 & 80\\
     II.11.28 & $1 + \frac{n \alpha}{1 - (n \alpha / 3)}$ & 20 & 100 \%  & 100 & 100 & 80\\
     I.6.2a & $\frac{\exp(-\theta^2/2)}{\sqrt{2\pi}}$ & 0& 100 \% & 100 & 100& 100\\
    II.35.18 & $\frac{n_0}{\exp(\mathrm{mom} B / (k_b T)) + \exp(-\mathrm{mom} B / (k_b T))}$ & 80 & 100 \% & 40 & 100 & 80 \\
    \hline
    Total &  & 57.7 \% & 100\% & 75.8 \% & 90.8 \% & 93.7 \% \\
    \hline
    \end{tabular}
    \caption{Easy targets for our main algorithm are those solved in just a few iterations, typically within a few minutes. The last column represents the main run's performance on noisy data with a 0.1 noise level. The hit rate in this case is defined as explained in Section 3.11}.
    \label{tab:easy2}
\end{table}
Easy targets are those that are solved in a very short amount of time, typically requiring fewer than 10,000 individuals to be evaluated, that is less than a few dozen genetic iterations. In fact, many of these targets are often identified right from the initial random pool, i.e., at iteration 0. 

As discussed earlier in Section 3.4, we highlight that targets III.10.19 and I.8.14 in the easy table, in particular, greatly benefit from dimensional analysis (compare the hit rates between the main run and Ab\_0), while equation I.9.18 in Eq.~(\ref{1918}) requires both dimensional analysis and an extended vocabulary to be successfully recovered.
\newpage
\begin{table}[ht]
    \centering
    \begin{tabular}{|l|c|c|>{\columncolor[gray]{0.9}}c|c|c|c|}
    \hline
    \multicolumn{7}{|c|}{\textbf{20 Medium Targets}} \\ 
    \hline
    \multicolumn{2}{|c|}{\textbf{Equation Details}} & \multicolumn{5}{c|}{\textbf{Hit Rates in \%}} \\
    \hline
    \textbf{Eq. Id.} & \textbf{Formula} & \textbf{PySR} & \textbf{Main} & \textbf{Abl 0} & \textbf{Abl 1} & \textbf{w. 0.1 noise} \\
    \hline
        III.8.54 & $\sin^2\left(\frac{E_n t}{h / (2 \pi)}\right)$ & 80 & 100 \% & 80 & 80 & 100\\
        I.15.3x & $\frac{x - u t}{\sqrt{1 - u^2 / c^2}}$ & 0 & 80 \% & 40 & 80 & 20\\
        I.30.5 & $\arcsin(\lambda / (n d))$ & 100 & 80 \%  & 60 & 100 & 60\\
         $\mathrm{B}_{19}$ & $-\frac{1}{8 \pi G} \left( \frac{c^4 k_f}{r^2} + H_G^2 c^2 (1 - 2\alpha) \right)$ & 40 & 100 \% & 60 & 100 & 80\\
        I.15.3t & $\frac{t - u x / c^2}{\sqrt{1 - u^2 / c^2}}$ & 20 & 100 \%  & 20 & 100 & 40\\
        $\mathrm{B}_7$ & $\sqrt{\frac{8 \pi G \rho}{3} - \frac{\alpha c^2}{d^2}}$ & 60 & 100 \%  & 20 & 40 & 20 \\
        $\mathrm{B}_{17}$ & $\frac{1}{2 m} \left( p^2 + m^2 \omega^2 x^2 \left( 1 + \frac{\alpha x}{y} \right) \right)$ & 100 & 100 \% & 20 & 60 & 60\\
        $\mathrm{B}_8$ & $\frac{E_n}{1 + \frac{E_n}{m c^2} (1 - \cos\theta)}$ & 60 & 100 \% & 40 & 100 & 100 \\
        I.6.2 & $\frac{\exp(- (\theta / \sigma)^2 / 2)}{\sqrt{2\pi} \sigma}$ & 20 & 100 \% & 100 & 100 & 20\\
        I.30.3 & $I_0 \frac{\sin(n \theta / 2)^2}{\sin(\theta / 2)^2}$ & 100 & 100 \%  & 80 & 80 & 80\\
        I.37.4 & $I_1 + I_2 + 2 \sqrt{I_1 I_2} \cos(\delta)$ & 60 & 100 \%  & 100 & 100 & 60\\
        $\mathrm{B}_6$ & $\sqrt{1 + \frac{2 \varepsilon^2 E_n L^2}{m (Z_1 Z_2 q^2)^2}}$ & 0 & 90 \%  & 0 & 0  & 40 \\
        I.16.6 & $\frac{u + v}{1 + u v / c^2}$ & 60 & 90 \% & 60 & 100 & 20\\
        $\mathrm{B}_4$ & $\sqrt{\frac{2}{m} \left( E_n - U - \frac{L^2}{2 m r^2} \right)}$ & 60 & 100 \% & 40 & 60 & 0\\
        I.26.2 & $\arcsin(n \sin \theta_2)$ & 0 & 100 \% & 100 & 60 & 60 \\
        $\mathrm{B}_1$ & $\left(\frac{Z_1 Z_2 \alpha \hbar c}{4 E_n \sin^2(\theta/2)}\right)^2$ & 0 & 100 \% & 20 & 80 & 0\\
        $\mathrm{B}_2$ & $\frac{m k_G}{L^2} \left( 1 + \sqrt{1 + \frac{2 E_n L^2}{m k_G^2}} \cos(\theta_1 - \theta_2) \right)$ & 0 & 100 \% & 0 & 0 & 20\\
        II.6.15a & $\frac{p_d}{4 \pi \varepsilon} \frac{3 z}{r^5} \sqrt{x^2 + y^2}$ & 0 & 80 \% & 60 & 60 & 60\\
        $\mathrm{B}_{14}$ & $E_f \cos\theta \left( -r + \frac{d^3}{r^2} \frac{(\alpha - 1)}{\alpha + 2} \right)$ & 20 & 80 \% & 20 & 80 & 0\\
        $\mathrm{B}_3$ & $d \frac{1 - \alpha^2}{1 + \alpha \cos(\theta_1 - \theta_2)}$ & 0 & 90 \% & 60 & 60 & 20\\
        \hline
        Total &  & 39 \% & 94.5\% & 49\%  & 72 \% & 43 \% \\
        \hline
    \end{tabular}
    \caption{More difficult targets, sorted by the average number of individuals seen (in ascending order, ranging from around 20,000 to 250,000). The last equations are typically found, but they may require several hours of runtime.}
    \label{tab:medium}
\end{table}
\newpage
\renewcommand{\arraystretch}{1.25} 
\begin{table}[ht]
    \centering
    \begin{tabular}{|l|c|c|>{\columncolor[gray]{0.9}}c|c|c|c|}
    \hline
    \multicolumn{7}{|c|}{\textbf{14 Difficult Targets}} \\ 
    \hline
    \multicolumn{2}{|c|}{\textbf{Equation Details}} & \multicolumn{5}{c|}{\textbf{Hit Rates in \%}} \\
    \hline
    \textbf{Eq. Id.} & \textbf{Formula} & \textbf{PySR} & \textbf{Main} & \textbf{Abl 0} & \textbf{Abl 1} & \textbf{w. 0.1 noise} \\
    \hline
        III.4.32 & $\frac{1}{\exp((h/(2 \pi)) \omega/(k_b T)) - 1}$ & 0 & 30 \% & 40 & 80 & 40\\
        III.4.33 & $\frac{(h/(2 \pi)) \omega}{\exp((h/(2 \pi)) \omega/(k_b T)) - 1}$ & 0 & 20 \%  & 0 & 0 & 40\\
        I.6.2b & $\frac{\exp(- ((\theta - \theta_1) / \sigma)^2 / 2)}{\sqrt{2\pi} \sigma}$ & 0 & 40 \% & 40 & 20& 0 \\
        II.24.17 & $\sqrt{\frac{\omega^2}{c^2} - \frac{\pi^2}{d^2}}$ & 0 & 70 \%  & 60 & 100& 60\\
        I.41.16 & $\frac{h}{2 \pi} \frac{\omega^3}{\pi^2 c^2 (\exp((h/(2 \pi)) \omega/(k_b T)) - 1)}$ & 0 & 40 \%  & 0 & 60& 40\\
        $\mathrm{B}_{12}$ & $\frac{q}{4 \pi \varepsilon y^2} \left( 4 \pi \varepsilon \mathrm{Volt} \cdot d - \frac{q d y^3}{(y^2 - d^2)^2} \right)$ & 0 & 50 \% & 20 & 60& 0\\
        $\mathrm{B}_{15}$ & $\frac{\sqrt{1 - v^2/c^2} \cdot \omega}{1 + (v/c) \cos\theta}$ & 0 & 70 \% & 20 & 80 & 20\\
        $\mathrm{B}_{16}$ & $\sqrt{(p - q \mathbf{A})^2 c^2 + m^2 c^4} + q \mathrm{Volt}$ & 0 & 60 \% & 0 & 20 & 20\\
        $\mathrm{B}_{13}$ & $\frac{1}{4 \pi \varepsilon} \frac{q}{\sqrt{r^2 + d^2 - 2 r d \cos\alpha}}$ & 0 & 40 \%  & 0 & 20 & 20\\
        I.29.16 & $\sqrt{x_1^2 + x_2^2 - 2 x_1 x_2 \cos(\theta_1 - \theta_2)}$ & 0 & 10 \% & 0 & 0& 0\\
        III.9.52 & $\frac{(p_d E_f t / (h / (2 \pi))) \sin^2((\omega - \omega_0) t / 2)}{((\omega - \omega_0) t / 2)^2}$ & 0 & 0 \% & 0 & 0& 0\\
        $\mathrm{B}_{20}$ & $\frac{1}{4 \pi} \frac{\alpha^2 h^2}{m^2 c^2} \left( \frac{\omega_0}{\omega} \right)^2 \left( \frac{\omega_0}{\omega} + \frac{\omega}{\omega_0} - \sin^2\beta \right)$ & 0 & 0 \% & 0 & 0& 0\\
        $\mathrm{B}_{10}$ & $\arccos \left( \frac{\cos\theta_2 - v/c}{1 - (v/c) \cos\theta_2} \right)$ & 0 & 0 \% & 0 & 0& 0\\
        $\mathrm{B}_{11}$ & $I_0 \left( \frac{\sin(\alpha/2) \sin(n \delta/2)}{(\alpha/2) \sin(\delta/2)} \right)^2$ & 0 & 0 \% & 0 & 0& 0\\
        \hline
        Total & & 0 \% & 30.7 \% &  12.8\% & 31.4\%   & 18.4 \%  \\
        \hline
    \end{tabular}
    \caption{Difficult targets. The first five are challenging because early stopping is often triggered on formulas that are inexact but very close to numerical precision. The next four are difficult and may sometimes lead to false positives. The remaining five seem to be  inherently difficult. These targets fail even after 6-hour runs, although good accuracy is achieved.}
    \label{tab:hard}
\end{table}

\acks{This work is supported by Paris-Cité University. LLM models have been used to improve spelling and syntax.}

\newpage 

\newpage 

\appendix

\section{Random Tree Generator}
Our goal is to generate a large number of valid and diverse trees rapidly while minimizing failure rates (i.e., cases where a tree fails to terminate within the allowed number of nodes). To achieve this, we employ a structured tree-generation approach that ensures the generated equations have dimensions matching a given target dimension, given a set of leaf variables \((x_i, A)\) and their corresponding dimensions.

\paragraph{Precomputing Dimensionally Valid Monomials}
We first construct a dictionary of all the dimensions that can be generated by monomials of the form $\prod x_i^{q_i}, \, q_i \in [0, 1, \ldots, p]$ for some integer \( p \). This will be used to split dimensions when traversing down the tree after encountering a 
$(\times, \div, **)$ operator.

\paragraph{Tree Initialization and Expansion Strategy} Tree generation starts with a root node that initially carries only the target dimension. We maintain both a queue of nodes to expand, initialized with the root node, and a counter for the total number of expanded nodes to enforce a maximum tree size constraint. The expansion process then follows these rules:
\begin{enumerate}
    \item Termination Conditions: The process stops when either the maximum allowed number of nodes is reached, or when there are no more nodes left to expand.
    \item Expanding a Node: Each node is dequeued and expanded by:
    \begin{itemize}
        \item Choosing its arity.
        \item Selecting an appropriate symbol given the arity.
        \item Assigning dimensions to child nodes, if not already constrained (this is required only for symbols $(\times, \div, **)$.
    \end{itemize}
\end{enumerate}

\paragraph{Choosing Node Arity}
A node’s arity is determined probabilistically, but subject to dimensional constraints. Binary arity is always possible, while unary arity is allowed if the node is dimensionless, or if square root is permitted, respecting a constraint on nested unaries. On the other hand, leaf nodes are possible if at least one variable in the vocabulary exactly matches the node’s dimension. Binaries and leaves are favored over unaries.

\paragraph{Selecting Operators and Symbols}
Once arity is determined, we assign an operator, applying again non-uniform probability distributions:
\begin{itemize}
    \item \textbf{Binary operators (\( +, -, \times, \div \))}:
    \begin{itemize}
        \item Multiplication (\(\times\)) is favored over division (\(\div\)), which in turn is favored over addition/subtraction (\(+, -\)). Power may also be chosen if dimension is carefully matched.
        \item This bias prevents excessive growth in tree size, as repeated addition and subtraction tend to make trees expand uncontrollably before reaching the node limit.
    \end{itemize}
    \item \textbf{Unary operators and leaf nodes} are chosen randomly.
\end{itemize}

\paragraph{Ensuring Dimensional Consistency}
If a node has two children and is assigned \( \times \) or \( \div \), the children’s dimensions must satisfy:
\[
\text{dim}(\text{left}) \pm \text{dim}(\text{right}) = \text{dim}(\text{parent})
\]
where \( + \) corresponds to multiplication and \( - \) to division. We resolve this constraint using the precomputed dictionary of dimensions via a basic two-sum or two-differences algorithm. If the dimension cannot be split, the algorithm stops and we have failed to generate a consistent tree.

\paragraph{Enhancing Diversity in Generated Trees}
To maximize diversity and reduce failure cases where trees do not terminate we :
\begin{itemize}
    \item Resample hyperparameters (such as \( p \), probability distributions for arities and operators) for each new trial.
    \item Attempt multiple strategies:
    \begin{itemize}
        \item We try direct tree generation, or
        \item We start by splitting, if possible, the target dimension as : a numerator dimension divided by a denominator dimension (following the lookup table) and then attempt to build both numerator and denominator tree directly, or
        \item We start by prefactorizing the target dimension as a function of the leaves variables (this is always possible), and then attempt at generating a dimensionless tree only.
    \end{itemize}
\end{itemize}
When all of this is implemented, we end up with an efficient tree generation process with a sufficiently low failure rate and a diverse initial 'random' pool of trees. However, the resulting tree lengths are not uniformly distributed across the number of nodes. In practice, our algorithm has a clear tendency to overproduce smaller trees compared to longer ones, which is not a desirable feature. This, however, is a minor issue, as crossovers in the QD grid will naturally generate longer expressions during the subsequent genetic iterations.

\vskip 0.2in
\bibliography{bib_no_url}

\end{document}